\documentclass[conference]{IEEEtran}
\IEEEoverridecommandlockouts
\usepackage{soul}
\usepackage{cite}
\usepackage{amsmath,amssymb,amsfonts}
\usepackage{algorithmic}
\usepackage{graphicx}
\usepackage{textcomp}
\usepackage{xcolor}
\usepackage{float}
\def\BibTeX{{\rm B\kern-.05em{\sc i\kern-.025em b}\kern-.08em
    T\kern-.1667em\lower.7ex\hbox{E}\kern-.125emX}}
\begin{document}

\title{VLMs-in-the-Wild: Bridging the Gap Between Academic Benchmarks and Enterprise Reality\\
}

\author{\IEEEauthorblockN{Srihari Bandraupalli}
\IEEEauthorblockA{\textit{Intern} \\
\textit{Sprinklr}\\
Gurgaon, India \\
}
\and
\IEEEauthorblockN{Mihir Gajera}
\IEEEauthorblockA{\textit{AI Team} \\
\textit{Sprinklr}\\
Gurgaon, India \\
}
\and
\IEEEauthorblockN{Ali Saifee}
\IEEEauthorblockA{\textit{AI Team} \\
\textit{Sprinklr}\\
Gurgaon, India \\
}
\and
\IEEEauthorblockN{Anupam Purwar\IEEEauthorrefmark{1}}
\IEEEauthorblockA{\textit{AI Team} \\
\textit{Sprinklr}\\
Gurgaon, India \\
}
\thanks{\IEEEauthorrefmark{1}Corresponding Author: Anupam Purwar (e-mail: anupam.aiml@gmail.com, https://anupam-purwar.github.io/page/)}
}
\maketitle

\begin{abstract}
Open-source Vision-Language Models show immense promise for enterprise applications, yet a critical disconnect exists between academic evaluation and enterprise deployment requirements. Current benchmarks rely heavily on multiple-choice questions and synthetic data, failing to capture the complexity of real-world business applications like social media content analysis. This paper introduces VLM-in-the-Wild (ViLD), a comprehensive framework to bridge this gap by evaluating VLMs on operational enterprise requirements. We define ten business-critical tasks: logo detection, OCR, object detection, human presence and demographic analysis, human activity and appearance analysis, scene detection, camera perspective and media quality assessment, dominant colors, comprehensive description, and NSFW detection. To this framework, we bring an innovative BlockWeaver Algorithm that solves the challenging problem of comparing unordered, variably-grouped OCR outputs from VLMs without relying on embeddings or LLMs, achieving remarkable speed and reliability. Besides, ViLD's methodology avoids traditional bounding boxes, which are ill-suited for generative VLMs, in favour of a novel spatial-temporal grid system that captures localisation information effectively for both images and videos. To demonstrate efficacy of ViLD, we constructed a new benchmark dataset of 7,500 diverse samples, carefully stratified from a corpus of one million real-world images and videos. ViLD provides actionable insights by combining semantic matching (both embedding-based and LLM-as-a-judge approaches), traditional metrics, and novel methods to measure the completeness and faithfulness of descriptive outputs. By benchmarking leading open-source VLMs (Qwen, MIMO, and InternVL) against a powerful proprietary baseline as per ViLD framework, we provide one of the first industry-grounded, task-driven assessment of VLMs capabilities, offering actionable insights for their deployment in enterprise environments.
\end{abstract}


\section{Introduction}
Vision-Language Models (VLMs) have fundamentally transformed the landscape of artificial intelligence, enabling systems to understand and reason about visual content through natural language. This progress holds immense promise for enterprise domains that require rich, multimodal analysis: from social media analysis and brand intelligence to automated market research and customer analytics. Yet, despite rapid progress, a critical disconnect between how these models are evaluated in academic settings and the practical requirements of real-world business deployment. The architectural evolution of VLMs has fundamentally changed both their capabilities and evaluation requirements. 

Early contrastive models like \textbf{CLIP}\cite{clip} and \textbf{ALIGN}\cite{align} established the foundation by learning joint embeddings between images and text through contrastive learning on large-scale paired datasets. These models excelled at zero-shot classification tasks and were typically evaluated using accuracy metrics on curated datasets like ImageNet, where their structured outputs (class probabilities) aligned well with traditional evaluation frameworks. The next generation of instruction-tuned “fusion” models, such as \textbf{BLIP-2}\cite{blip2} and \textbf{InstructBLIP}\cite{instructblip}, moved towards richer multimodal reasoning. These systems enabled conversational multimodal outputs and the ability to follow natural language instructions, prompting the first wave of VQA-style (Visual Question Answering) benchmarks.

The latest end-to-end generative VLMs represent a paradigm shift toward native text generation. Models like LLaVA\cite{llava}, MiniGPT-4\cite{minigpt4}, InternVL\cite{internvl} directly project visual features into large language models' input spaces, making rich descriptive text generation their native capability rather than an afterthought. This architectural evolution from discriminative outputs (discrete class labels, bounding boxes, or sequences) to generative outputs (rich descriptive paragraphs capable of conveying nuanced, context-aware insights) creates the core evaluation challenge that current benchmarks fail to address. 

Although there have been significant efforts to benchmark VLMs, the overwhelming majority of existing benchmarks are designed for academic contexts and utilise task formats out of sync with enterprise requirements as discussed in Table \ref{tab:vlm_limitations}.

\begin{table*}[htbp]
\caption{Limitations of Common VLM Benchmarks in Enterprise Settings}
\begin{center}
\begin{tabular}{|c|c|c|c|}
\hline
\textbf{Benchmark Type} & \textbf{Representative Examples} & \textbf{Evaluation Format} & \textbf{Key Limitation for Enterprise} \\
\hline
Reasoning & MMMU, ScienceQA, MathVista & Multiple-choice, VQA & Academic focus, misses generative/business reasoning \\
\hline
Perception/Hallucination & MME, MMBench, POPE & Binary/MC, polling & Cannot assess open-ended, descriptive, or structured outputs \\
\hline
Domain-specific & TextVQA, DocVQA, ChartQA & QA in narrow verticals & Siloed tasks, little cross-modal or operational integration \\
\hline
Video/Temporal & Video-MME & MC/VQA & Limited free-form outputs \\
\hline
\end{tabular}
\label{tab:vlm_limitations}
\end{center}
\end{table*}

\textbf{Reasoning-centric benchmarks} such as MMMU\cite{mmmu}, ScienceQA\cite{scienceqa}, and MathVista\cite{mathvista} consist primarily of multiple-choice (MC) or structured visual question answering (VQA) tasks, often drawn from synthetic or academic content. These benchmarks excel at measuring abstract cognitive abilities, but their multiple-choice format, while enabling precise evaluation, cannot assess a model's ability to generate the rich, structured descriptions that enterprise applications require.

\textbf{Perception and hallucination benchmarks}, including MME\cite{mme}, MMBench\cite{mmbench}, and POPE\cite{pope}, focus on basic object and attribute recognition, hallucination detection by reducing complex visual understanding to yes/no questions. Such diagnostics are valuable for model development but provide little evidence of a system’s performance on open-ended, structured outputs such as listing all brands, reading and grouping textual elements, or localising and contextualising brand appearances in a natural scene.

\textbf{Domain-specific benchmarks}, like AI2D\cite{ai2d}, TextVQA\cite{textvqa}, DocVQA\cite{docvqa}, ChartQA\cite{chartqa} make critical contributions within isolated verticals (OCR, document QA, chart reasoning) but reinforce a fragmented “pipeline” approach. This siloing falls short of capturing the integrative, multi-faceted nature of real business streams, where VLMs are tasked to synthesise insights across text, objects, demographics, and brand contexts within a single input.

\textbf{Video understanding benchmarks} such as Video-MME\cite{videomme} target temporal or dialogue-based reasoning but remain largely limited by continued reliance on multiple-choice or binary QA formats.

Our systematic analysis reveals a \textbf{three-fold gap} between current VLM evaluation and enterprise deployment needs. First, the \textbf{Task Gap} arises because existing benchmarks prioritise academic reasoning over operational intelligence tasks, as enterprise-critical needs are marginalised or missing. Second, the \textbf{Data Gap} stems from the fact that current datasets are too clean or too narrow to represent the chaotic real-world data with overlaid text, subtle brand placements, and multilingual content often common in enterprise applications. Third, the \textbf{Modality Gap} reflects how the shift from discriminative to generative VLMs has rendered traditional evaluation metrics obsolete. Bounding-box mAP assumes structured coordinate outputs, while character error rates assume ordered text strings. In contrast, generative VLMs produce free-form descriptions that may list detected elements in any order or use varied phrasings. Standard NLP metrics like BLEU and ROUGE are similarly inadequate, as they penalise valid synonyms and assume fixed word order. This mismatch leaves enterprises without reliable methods to assess the VLMs.

Bridging these gaps is both urgent and challenging. The prevailing enterprise reality is a “pipeline-of-pipelines” paradigm, where multiple siloed models (object detectors, OCR systems, custom brand detectors) must be individually maintained and integrated, leading to significant complexity and scaling overhead. VLMs offer a path toward unification, but this potential can only be realised if their performance on real-world enterprise tasks is evaluated using methods that account for practical deployment constraints and variability in output formats.

In response, we introduce \textbf{VLM-in-the-Wild}, a comprehensive evaluation framework designed from the ground up for enterprise VLM applications and deployment. 



Through this comprehensive framework, VLM-in-the-Wild provides a novel framework grounded in enterprise based evaluation of VLM capabilities. ViLD offers the data driven confidence across multiple tasks essential to deployment of VLMs in real-world business applications.

\section{Evaluation Dataset}

\begin{figure*}[h]
    \centering
    \includegraphics[width=\linewidth]{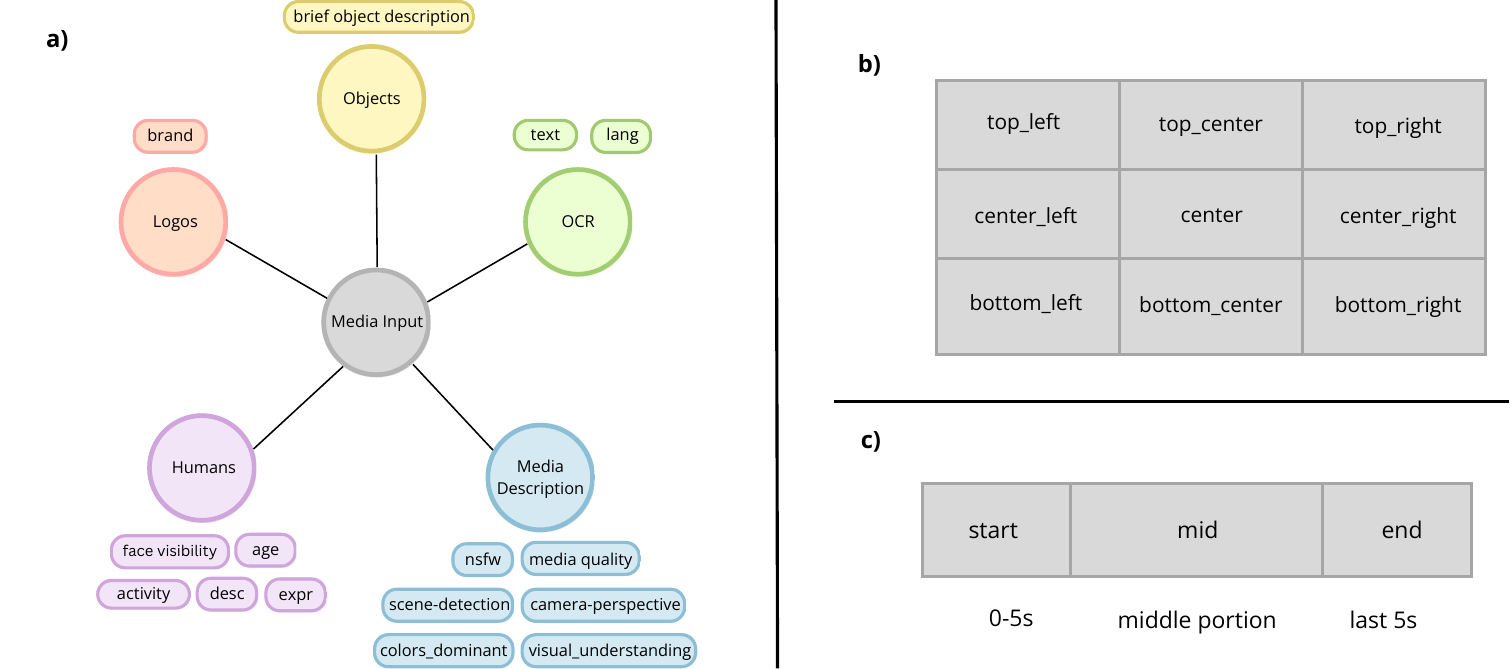}
    \caption{Overview of our multi-task annotation and localization framework. \textbf{(a)} The unified JSON schema for structured, multi-task annotation. \textbf{(b)} The standardized \(3 \times 3\) spatial grid used for localizing entities within images (e.g., \texttt{top-left}, \texttt{center}). \textbf{(c)} The temporal grid for segmenting videos into key intervals (\texttt{start}, \texttt{mid}, \texttt{end}, and \texttt{inter}) to track entity presence over time.}
    \label{fig:json_schema}
\end{figure*}

Robust enterprise evaluation of Vision-Language Models (VLMs)/LLMs demands a dataset that not only embodies the operational diversity and complexity of real business media, but is also structured to support nuanced, multi-task evaluation \cite{gautam2024opensourceLLMsEnterpriseRAG}. This section details the principles guiding our task selection, introduces our spatio-temporal localization grid, and presents our unified annotation schema alongside the dataset construction process.

\subsection{Task Selection Grounded in Enterprise Reality}
To ensure that our benchmark is genuinely reflective of enterprise needs, we undertook a rigorous survey of current industry deployments, internal requirements from large-scale enterprises, and forward-looking use cases that VLMs are expected to unlock\cite{VLMeval}. This process distilled ten business-critical vision-language tasks (as shown in Figure~\ref{fig:json_schema}), each corresponding to a common bottleneck or opportunity for automation in enterprise workflows:

\begin{itemize}
    \item \textbf{Logo Detection:} Identifies brand presence, which is critical for competitive analysis, brand monitoring, and measuring sponsorship ROI.
    
    \item \textbf{General Object Detection:} Pinpointing and describing key objects underpins product tracking, automated content tagging, and supply chain surveillance.
    
    \item \textbf{Optical Character Recognition (OCR):} Extracts textual information from images and videos, essential for understanding user-generated content, processing in-media advertisements, and flagging prohibited text.
    
    \item \textbf{Human Presence \& Demographic Analysis:} Detecting individuals, estimating age group, and expressions support targeted advertising, customer sentiment analysis, and safety compliance.
    
    \item \textbf{Human Activity \& Appearance Analysis:} Describes the actions, poses, and visual appearance of individuals, providing insight into consumer behaviour, human-object interactions, and demographic targeting.
    
    \item \textbf{Scene Detection:} Identifies specific environmental settings, lighting conditions, and atmospheric context, which is vital for event detection, trend analysis, and location-based insights.
    
    \item \textbf{Camera Perspective and Media Quality Assessment:} Determining viewpoint, camera movement, and technical quality guides media curation, selection, and automated editing processes.
    
    \item \textbf{Dominant Color Extraction:} Powers brand-aligned marketing analytics and visual search systems.
    
    \item \textbf{Comprehensive Description:} A capstone task that generates a detailed, integrated summary of the entire media, testing the model's ability to synthesize all other elements into a coherent narrative.
    
    \item \textbf{Content Moderation (NSFW Detection):} Mitigates risk and ensures compliance by detecting inappropriate or unsafe content.
\end{itemize}

Each task was selected based on a clear business imperative, grounded in a systematic review of one million samples from enterprise data streams.

\subsection{Spatio-Temporal Localization Grid}

Traditional computer vision models rely on pixel-level bounding boxes for evaluating localization performance\cite{yolo}. However, generative VLMs, which output textual descriptions rather than spatial coordinates, are fundamentally unsuited to such regression-based evaluation\cite{visualgrounding}. To address this challenge, we introduce a flexible \textbf{spatio-temporal grid} framework that captures essential localization information while accommodating both generative reasoning and the specificity required by enterprise applications.

For images, we employ a standardized \(3 \times 3\) spatial grid as shown in Figure~\ref{fig:json_schema}b, with positions such as \texttt{top-left}, \texttt{center}, and \texttt{bottom-right}. This granularity sufficiently addresses real enterprise requirements, such as logo placement for brand compliance, where determining approximate region is sufficient. The grid-based system aligns naturally with how VLMs typically describe object positions while providing the spatial specificity needed for business decision-making.

For videos, this approach is extended temporally as illustrated in Figure~\ref{fig:json_schema}. Media are segmented into three primary intervals: \texttt{start} (first 5 seconds), \texttt{mid} (middle portion), and \texttt{end} (last 5 seconds), with an additional \texttt{inter} label for elements that appear intermittently throughout the video. This temporal reference system allows models to track objects across video sequences and provide a list of their temporal presence without the complexity of frame-level or second-level positioning. Such fine-grained temporal tracking would be impractical given that objects can appear and disappear multiple times, making our interval-based approach both more robust and interpretable.

Collectively, this approach bridges the gap between human-interpretable, descriptive localization and the practical needs of enterprise VLM deployment---enabling fine-grained, multi-modal assessment without penalizing generative models for not predicting spatial coordinates.

\subsection{Unified Schema Multi-Task Annotation Schema}

To capture the rich, multi-faceted information required by our defined tasks in a structured and machine-readable format, we designed a unified JSON output schema. Our schema (visualized in Figure~\ref{fig:json_schema}a) provides distinct top-level fields for \texttt{objects}, \texttt{humans}, \texttt{logos}, \texttt{ocr}, and \texttt{media\_description}, tightly aligning with the task taxonomy. This modular structure facilitates task-specific evaluation and prevents a single monolithic description from becoming convoluted. Each entity includes grid- and (for video) temporal pointers (\texttt{pos}, \texttt{temp}), a \texttt{conf} confidence score, and relevant descriptive subfields.

Key design decisions are empirically grounded based on analysis of our source corpus. For OCR annotations, we encourage the grouping of spatially proximate text into larger blocks, reflecting human reading patterns. Video samples are capped at 3 minutes, while object and human detection are limited to the 10 and 5 most salient instances per sample, respectively---guided by corpus analysis covering 95\% of observed cases. Additional individuals are efficiently handled as groups. The schema's structured support for scene context, camera perspective, quality assessment, and comprehensive description enables both targeted and holistic evaluation.

\subsection{Dataset Curation and Statistics}

\begin{figure}[htbp]
\centering
\includegraphics[width=\linewidth]{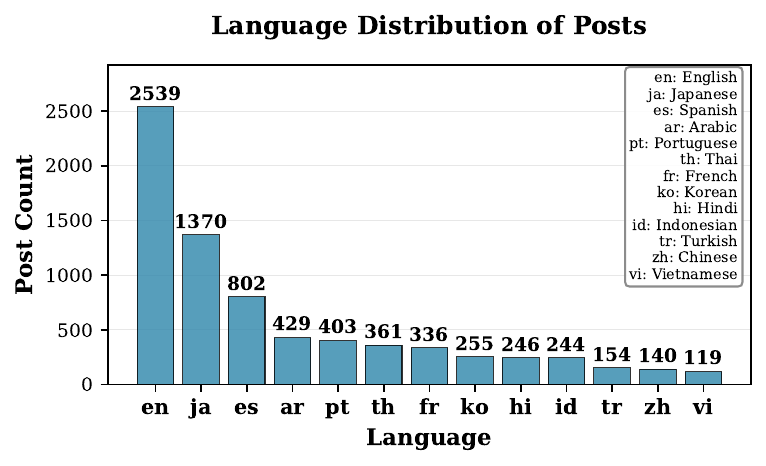}
\caption{Distribution of posts across different languages in the dataset.}
\label{fig:lang_distribution}
\end{figure}

The foundation of our evaluation corpus was a comprehensive collection of one million publicly available multimedia samples from large-scale online streams. This dataset reflects a broad spectrum of global events, marketing content, and organic user-generated media. \footnote{All data was collected in compliance with platform usage policies and is used strictly for non-commercial research purposes.}

To ensure statistical rigor, we employed a stratified sampling strategy: samples were balanced across language groups, content types (people-centric, text-centric, logo-rich), media quality, and prevalence of key enterprise-relevant phenomena. This stratification prevents evaluation overfitting to dominant content types and ensures that VLMs are challenged across the full range of real-world visual communication.

The final benchmark consists of around \textbf{7,500 media samples} (5,509 images and 1,889 videos), with an additional 500 NSFW samples for content moderation evaluation, totaling 7,898 samples. The dataset exhibits rich linguistic diversity across 13 distinct languages, with exact distribution counts shown in Figure~\ref{fig:lang_distribution}. This multilingual coverage ensures comprehensive evaluation across diverse textual contexts, including samples containing primarily symbols or language-indeterminate content.

Ground truth annotations conforming to our detailed JSON schema were generated using a powerful proprietary VLM (Gemini 2.5 Flash), guided by structured prompts tailored to our schema. While we acknowledge the limitations of an ``LLM-as-judge'' methodology for open-ended descriptive outputs, this represents current best practice for scalable, high-quality annotation of nuanced, multi-modal content, enabling consistent ground truth at a scale unattainable through manual human labeling \cite{LlmAsJudge} \cite{harbola2025knowslm}.

This dataset, by design and curation, establishes a new standard for comprehensive, enterprise-ready VLM evaluation, addressing the true spectrum of tasks and challenges encountered in real-world deployment.

\section{VLM-in-the-Wild Framework}
\begin{figure*}[t]
\centering
\includegraphics[width=\textwidth]{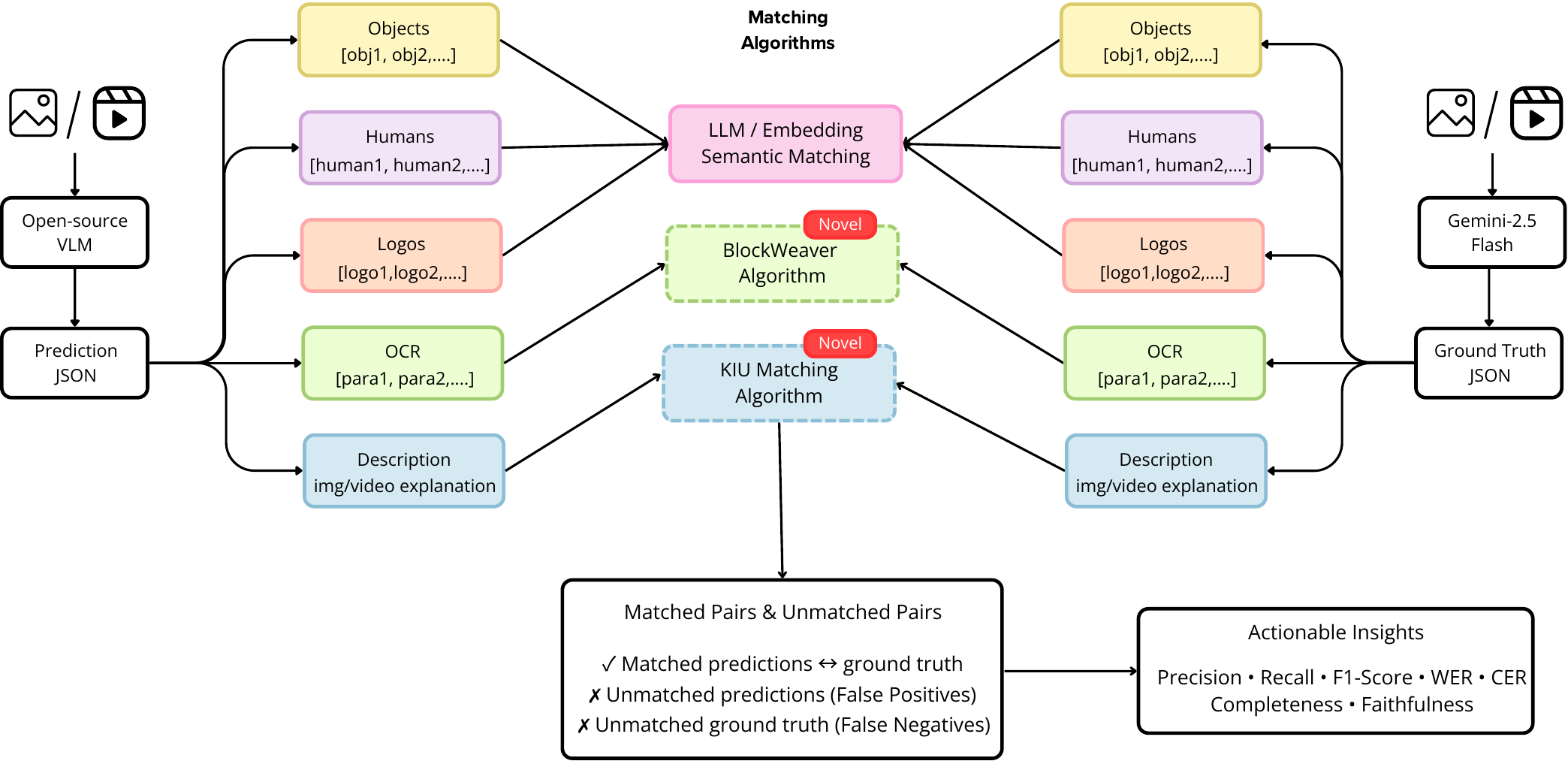}
\caption{Overview of the VLM-in-the-Wild evaluation framework used to benchmark vision-language models on real-world media data.}
\label{fig:evaluation-pipeline}
\end{figure*}

The core innovation of \textsc{VLM-in-the-Wild} lies in its comprehensive evaluation methodology that addresses the fundamental challenges of assessing generative Vision-Language Models in enterprise contexts. Unlike traditional computer vision evaluation that relies on structured outputs with fixed formats, generative VLMs produce free-form textual descriptions that vary in ordering, phrasing, and granularity. Our framework introduces novel techniques to robustly evaluate these outputs across multiple dimensions while maintaining the precision required for enterprise decision-making.

Figure~\ref{fig:evaluation-pipeline} illustrates our evaluation pipeline, which processes both predicted and ground truth outputs through specialized matching algorithms before computing actionable performance metrics. The framework handles five core evaluation domains: object detection, human analysis, logo recognition, OCR extraction, and comprehensive media description.

\subsection{The Entity Matching Challenge}

The evaluation of entity detection in generative VLMs presents a fundamental matching problem that distinguishes it from traditional computer vision tasks. While classical detectors output ordered lists of bounding boxes matched using IoU thresholds, generative VLMs produce unordered textual descriptions of varying granularity and specificity. This creates several critical evaluation challenges:

\textbf{Unordered Output Matching:} VLMs may describe detected entities in any sequence, making direct positional comparison impossible. A model might list ``smartphone, cup, notebook'' while the ground truth contains ``notebook, phone, coffee cup''---both represent perfectly valid detections requiring sophisticated matching algorithms.

\textbf{Cross-Category Ambiguity:} The same visual element may legitimately appear across different entity categories. For instance, an earring might be detected as a standalone ``object'' or described as part of a human's ``appearance'' analysis. Our framework must account for these valid cross-category representations without penalizing models for reasonable categorization choices.

\textbf{Granularity Variation:} VLMs exhibit varying levels of descriptive detail---one model might detect ``vehicle'' while another identifies ``red sedan,'' and a third specifies ``2019 Toyota Camry.'' All represent valid detections at different granularity levels, necessitating semantic rather than exact string matching.

\textbf{Semantic Equivalence:} VLMs may use semantically equivalent but lexically distinct terms. Detecting ``cellphone'' versus ``mobile phone'' or ``sofa'' versus ``couch'' should be recognized as identical detections, requiring evaluation methods that understand semantic relationships.

\subsection{Semantic Matching Methodology}

To address these challenges, we implement a comprehensive semantic matching system that provides robust correspondence between predicted and ground truth entities. Our approach employs carefully designed evaluation prompts tailored to capture the nuanced requirements of each entity type.

Extensive validation across approximately 10\% of our dataset reveals that LLM-based methods consistently outperform embedding-based approaches across all entity types. The LLM evaluation demonstrates superior handling of semantic nuances, contextual understanding, and strict visual presence requirements critical for enterprise applications. Embedding-based methods frequently produce false negatives through superficial semantic similarity that fails to reflect genuine visual detection accuracy.

The LLM-based matching employs task-specific prompts that emphasize strict visual presence criteria, distinguishing between explicit visual detection and implicit logical inference. For example, while ``microphone in hand'' clearly indicates a microphone's visual presence, ``person driving'' does not explicitly confirm a car's visual appearance. This emphasis ensures that our evaluation aligns with actual visual content analysis capabilities required in enterprise deployment scenarios.

Through this matching process, we establish correspondence between predicted and ground truth entities, enabling computation of standard detection metrics including precision, recall, and F1-score, as well as detailed attribute-level evaluation for spatial localization, temporal presence, and domain-specific characteristics.

\subsection{Object Detection Evaluation}

Object detection evaluation assesses whether VLMs can identify and describe key objects present in visual content. This capability underpins numerous enterprise applications including product tracking, automated content tagging, and supply chain surveillance.

Our evaluation employs a dual-prompt approach that separately assesses precision and recall. The precision evaluation determines whether each predicted object explicitly appears in the ground truth annotations, which include both dedicated object detections and objects mentioned within human activity descriptions. The recall evaluation checks whether each ground truth object is explicitly captured in the predicted statements.

This cross-category consideration reflects realistic output patterns of generative VLMs, where objects may be mentioned in multiple contexts. For instance, a ``basketball'' might appear both as a detected object and as part of a human activity description (``person holding basketball''). Our evaluation methodology credits both forms of detection while avoiding double-counting in precision calculations.

\subsection{Human Detection and Analysis}

Human detection and analysis evaluation assesses VLMs' ability to identify individuals and analyze their demographic characteristics, activities, and appearance attributes. This capability is crucial for targeted advertising, customer sentiment analysis, and behavioral analytics.

The evaluation process establishes correspondence between predicted and ground truth humans through a sophisticated matching algorithm that considers physical appearance characteristics and observable activities. We do not use spatial position as a primary matching criterion, as grid-based positions require separate evaluation as an attribute rather than a matching signal.

Following successful matching, we evaluate multiple human-specific attributes for each matched pair:

\begin{itemize}
    \item \textbf{Activity Accuracy}: Measures how well the predicted activity description matches the ground truth activity, including actions, gestures, poses, and overall behavior.
    \item \textbf{Physical Description Accuracy}: Assesses how well the predicted appearance description matches the ground truth description, including clothing, visible characteristics, and physical attributes.
    \item \textbf{Age Group Classification}: Evaluates exact match accuracy for predicted age categories against ground truth labels (``child'', ``teen'', ``adult'', ``elderly'').
    \item \textbf{Expression Recognition}: Measures exact match accuracy for facial expressions against ground truth labels (``happy'', ``sad'', ``neutral'', ``angry'').
    \item \textbf{Spatial Localization}: Computes Jaccard similarity between predicted and ground truth grid positions.
    \item \textbf{Temporal Localization}: For video content, measures Jaccard similarity between predicted and ground truth temporal intervals.
\end{itemize}

Detection performance is measured through precision, recall, and F1-score based on successful human matching, while attribute-level metrics provide detailed insights into specific capability dimensions.

\subsection{Logo Detection Evaluation}

Logo detection in VLMs is critical for brand monitoring, competitive analysis, and measurement of sponsorship exposure. Unlike object or human detection, logo identification requires careful handling of brand name variation, such as abbreviations, partial names, or colloquial terms.

The logo evaluation process handles brand name variations through LLM-based semantic matching. We prompt the LLM to recognize that brand variations, abbreviations, and alternative representations refer to the same commercial entity---for example, ``McDonald's,'' ``McDonalds,'' and ``McD's'' all represent valid detections of the same brand.

Our evaluation distinguishes between explicit visual presence (e.g., logo or branded product clearly visible) and references that are merely implied. Detections are only considered correct if there is explicit visual evidence for the brand, such as a logo, text, or branded product present in the content.

After establishing matches, we compute standard detection metrics---precision, recall, and F1-score---based on the number of correctly identified brands. For each matched logo, we further assess spatial localization (using Jaccard similarity on the spatial grid) and, for videos, temporal localization.

\subsection{OCR Evaluation: The BlockWeaver Algorithm}

\begin{figure*}[t]
\centering
\includegraphics[width=0.75\textwidth]{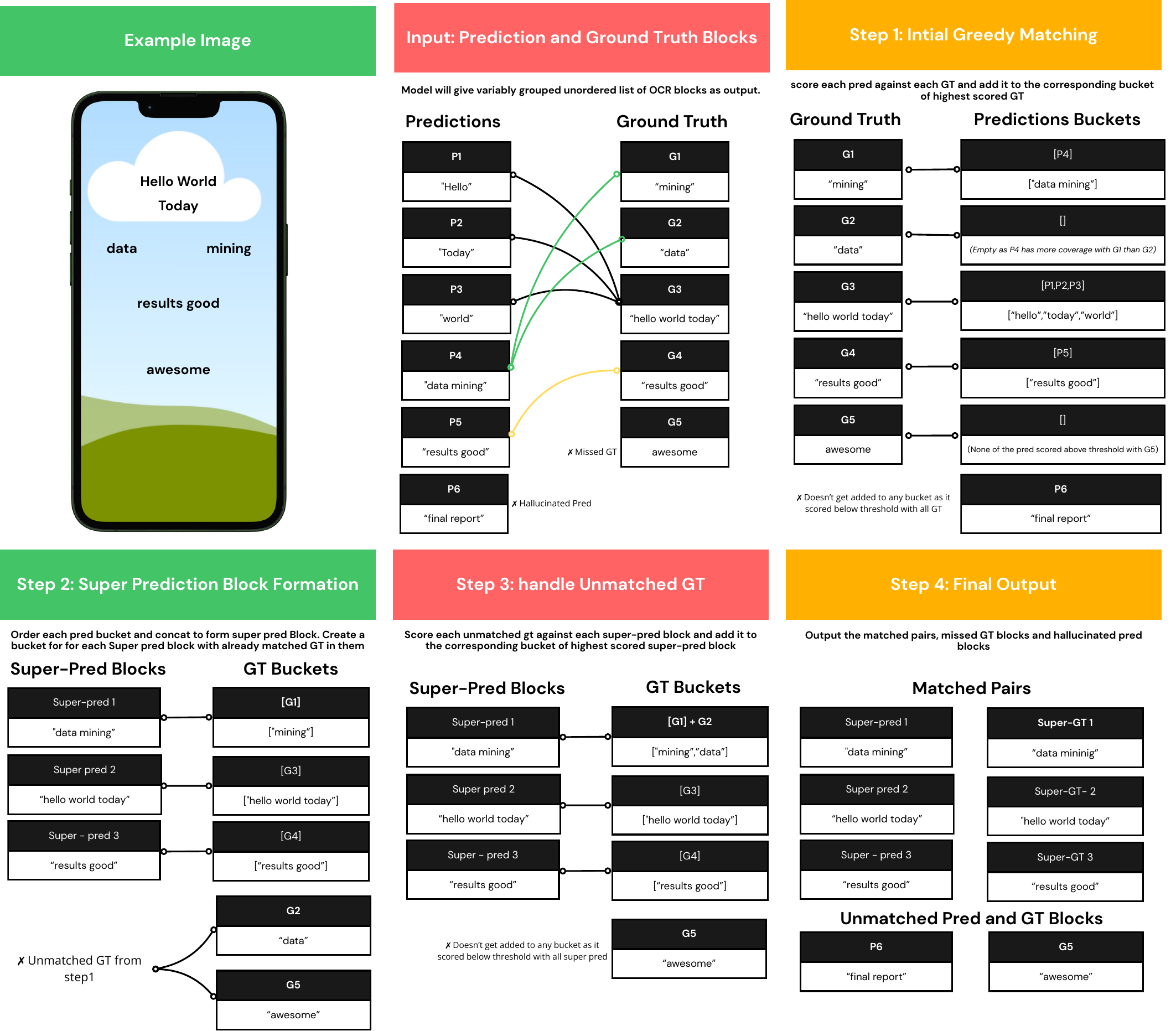}
\caption{BlockWeaver Algorithm Stepwise Illustration: (a) Example image and associated texts. (b) Sets of ground truth and prediction blocks, showing complex mappings. (c) Greedy assignment of predictions to GT blocks by coverage. (d) Super-prediction block composition via content ordering. (e) Bucketization and matched/unmatched GT assignment. (f) Output: matched super-block pairs plus unmatched blocks.}
\label{fig: Block-weaver}
\end{figure*}

Optical Character Recognition (OCR) evaluation in generative Vision-Language Models presents a fundamentally different challenge compared to traditional OCR systems. While conventional OCR models output structured text with spatial coordinates enabling direct comparison, generative VLMs produce unordered text blocks of varying granularity without reliable spatial information. This creates a complex matching problem where predicted text segments must be aligned with ground truth blocks despite arbitrary ordering, inconsistent segmentation strategies, and character-level OCR errors.

Standard text similarity metrics like BLEU, ROUGE, or character error rate fail catastrophically in this context because they assume ordered, consistently segmented text. A VLM might correctly identify all text content but segment it differently---for instance, outputting ``Hello World'' and ``Today'' as separate blocks while ground truth contains ``Hello World Today'' as a single block. Traditional metrics would severely penalize such outputs despite perfect content accuracy.

To address this challenge, we introduce the BlockWeaver algorithm, a novel approach designed specifically for matching variably-grouped OCR outputs without relying on spatial information, embedding similarity, or LLM calls. BlockWeaver transforms the complex many-to-many block correspondence problem into tractable one-to-one super-block pairs, enabling fair evaluation using standard metrics.

\subsubsection{Algorithm Design and Methodology}

Let $P = \{p_1, p_2, \ldots, p_n\}$ be the set of predicted text blocks and $G = \{g_1, g_2, \ldots, g_m\}$ be the set of ground truth text blocks. BlockWeaver operates through four sequential steps that progressively resolve block correspondences, as illustrated in Figure \ref{fig: Block-weaver}. The algorithm begins with these raw block sets and systematically creates super-blocks that capture both one-to-many and many-to-one relationships commonly observed in VLM outputs.

The foundation of BlockWeaver lies in its coverage-based scoring function, which measures textual overlap between any two blocks $A$ and $B$:

\begin{equation}
\text{Coverage-Score}(A, B) = \frac{\sum_{i} \text{length}(\text{matched\_substring}_i)}{\min(\text{length}(A), \text{length}(B))}
\end{equation}

This metric quantifies the proportion of the shorter text that finds correspondence in the longer text, providing robust similarity assessment despite character-level OCR errors and minor textual variations.

\textbf{Step 1: Initial Greedy Assignment} begins the matching process by assigning each prediction block to its highest-scoring ground truth block, provided the coverage score exceeds a threshold $\tau = 0.30$. As shown in Figure \ref{fig: Block-weaver}c, this creates initial correspondences where multiple prediction blocks may be assigned to the same ground truth block, while some blocks remain unmatched if no sufficiently strong correspondence exists.

For each prediction block $p_i \in P$, we compute:
\begin{equation}
\text{best\_gt}(p_i) = \arg\max_{g_j \in G} \text{Coverage-Score}(p_i, g_j)
\end{equation}

The assignment $p_i \rightarrow g_j$ is made only if $\text{Coverage-Score}(p_i, g_j) > \tau$, otherwise $p_i$ remains unmatched for later processing.

\textbf{Step 2: Super-Prediction-Block Formation} addresses the common scenario where multiple prediction blocks correspond to a single ground truth block. For each ground truth block $g_j$ with multiple assigned predictions $\{p_{i_1}, p_{i_2}, \ldots, p_{i_k}\} \subseteq P$, BlockWeaver orders these predictions based on their substring match positions within $g_j$ and concatenates them to form a super-prediction-block $SP_j$ (Figure \ref{fig: Block-weaver}d).

The ordering mechanism computes the weighted average position of matched substrings for each prediction block within the corresponding ground truth text. This ensures that super-blocks maintain logical text flow---for instance, predictions containing the beginning, middle, and end of a sentence are concatenated in the correct sequence rather than arbitrary order.

\textbf{Step 3: Unmatched Ground Truth Resolution} creates buckets $B_k$ for each super-prediction-block $SP_k$ and initially populates them with their already-matched ground truth blocks. Let $G_{\text{unmatched}}$ be the set of ground truth blocks not assigned in Step 1. The algorithm then processes each $g_j \in G_{\text{unmatched}}$, assigning it to the bucket corresponding to the super-prediction-block with the highest coverage score:

\begin{equation}
\text{best\_bucket}(g_j) = \arg\max_{SP_k} \text{Coverage-Score}(g_j, SP_k)
\end{equation}

The assignment is made only if the coverage score exceeds threshold $\tau$, as illustrated in Figure \ref{fig: Block-weaver}e. This step captures many-to-one relationships where multiple ground truth blocks correspond to a single prediction output.

\textbf{Step 4: Super-GT-Block Formation} completes the matching process by ordering ground truth blocks within each bucket $B_k$ and concatenating them to form super-GT-blocks $SG_k$. The final output consists of matched super-block pairs $\{(SP_k, SG_k)\}$ alongside sets $P_{\text{unmatched}}$ and $G_{\text{unmatched}}$ containing blocks that remained unmatched throughout the process (Figure \ref{fig: Block-weaver}f).

\subsubsection{Algorithm Characteristics and Limitations}

BlockWeaver successfully handles the most common OCR output variations observed in practice, including one-to-many relationships (single ground truth block matching multiple predictions), many-to-one relationships (multiple ground truth blocks matching single predictions), and standard one-to-one correspondences. The algorithm achieves $O(n \times m)$ computational complexity, enabling efficient evaluation across large datasets without requiring computationally expensive embedding calculations or LLM calls.

However, BlockWeaver operates under the fundamental assumption that VLMs produce reasonably ordered text output. The algorithm may fail in extreme edge cases where models exhibit highly disordered text segmentation---for instance, outputting the first and third words of a sentence as one block while separating the middle word. While such pathological cases are rare in practice, they represent the theoretical limitations of any matching algorithm operating without reliable spatial information.

Importantly, BlockWeaver offers the most comprehensive solution currently available for handling variable OCR block matching without spatial coordinates. Traditional approaches either require bounding box information for spatial alignment or rely on embedding-based similarity that fails to capture the precise textual accuracy requirements of OCR evaluation. Alternative solutions using LLM-as-a-judge approaches, while potentially more flexible, introduce computational overhead and evaluation inconsistency that make them impractical for large-scale benchmarking.

\subsubsection{Evaluation Integration}

Following super-block formation, BlockWeaver enables the application of standard OCR evaluation metrics—including character error rate, word error rate, and edit distance—on the matched pairs \(\{(SP_k, SG_k)\}\). In addition to these aligned pairs, the algorithm explicitly returns the set of unmatched prediction and ground truth blocks, allowing for more nuanced downstream analysis.

Crucially, by concatenating all super-ground-truth and super-prediction blocks, along with their unmatched counterparts, we construct complete predicted and reference texts for character-level evaluation. This enables computation of the total number of matched and unmatched characters, from which we derive a comprehensive character-level F1 score. This aggregated F1 metric, presented in Table~\ref{tab:overall_results}, captures the overall fidelity of the model’s OCR output while remaining robust to segmentation and ordering differences.

\subsection{Media Description Evaluation}

Media description evaluation addresses the critical enterprise need for comprehensive yet accurate content summarization, supporting applications from automated content tagging to accessibility compliance and regulatory documentation. Unlike discrete entity detection tasks, comprehensive media descriptions require assessment of both completeness (whether all key information is captured) and faithfulness (whether reported information is accurate), presenting unique evaluation challenges for generative VLMs.

Traditional text similarity metrics such as BLEU, ROUGE, or embedding cosine similarity fail to adequately capture the semantic richness and factual accuracy required for enterprise media description tasks. These metrics struggle with paraphrasing, synonym usage, and cannot distinguish between missing information (completeness issues) and hallucinated content (faithfulness issues), both critical concerns for enterprise deployment.

While direct LLM-as-a-judge evaluation of entire descriptions could provide holistic quality scores, such approaches yield limited actionable insights. A single aggregate score cannot identify specific types of omissions, hallucinations, or systematic model weaknesses, making it difficult for enterprise teams to understand model limitations or guide targeted improvements for specific deployment scenarios.

\subsubsection{The Key Information Units (KIU) Framework}

To address these limitations, we introduce the Key Information Units (KIU) matching algorithm, a novel approach that decomposes descriptive content into independently verifiable factual statements before performing semantic matching. The KIU framework operates through a two-stage process: first parsing both predicted and ground truth descriptions into atomic, self-contained information units, then matching these units using semantic similarity assessment.

The decomposition process addresses critical evaluation challenges through \textbf{granularity consistency} (grouping related attributes like ``a large, vintage, brown leather armchair'' into single units), \textbf{self-containment} requirements (eliminating ambiguous pronoun references), and \textbf{objective focus} (filtering subjective language that cannot be reliably verified).

This granular decomposition enables fine-grained analysis of model performance, identifying specific categories of missed information (temporal details, spatial relationships, object attributes) and common hallucination patterns (color misattribution, activity misidentification), providing actionable insights for model selection and deployment optimization.

\subsubsection{KIU Extraction and Matching}

The KIU extraction employs specialized prompts for consistent decomposition of complex descriptive text into verifiable factual units. Following extraction, our matching algorithm uses LLM-based semantic comparison to establish correspondence between predicted and ground truth information units, demonstrating superior performance compared to embedding-based alternatives in handling nuanced semantic relationships.

The matching process considers multiple forms of semantic equivalence including paraphrasing (``the woman is smiling'' $\leftrightarrow$ ``a woman shows a happy expression''), synonym usage, and conceptual overlap. The matching supports many-to-many relationships, reflecting how equivalent information may be expressed with different levels of detail across model outputs.
\begin{table*}[htbp]
\caption{Enterprise task-level performance across models for images and videos}
\begin{center}
\begin{tabular}{|c|c|c|c|c|c|c|}
\hline
\textbf{Model/Media} & \textbf{\textit{Reliability}} & \textbf{\textit{Object F1}} & \textbf{\textit{Human F1}} & \textbf{\textit{Logo F1}} & \textbf{\textit{OCR F1}} & \textbf{\textit{Media F1}} \\
\hline
\multicolumn{7}{|c|}{\textbf{Images}} \\
\hline
Qwen2.5-VL-7B & 0.97 & 0.39 & 0.78 & 0.54 & 0.75 & 0.70 \\
MIMO-SFT-7B & 0.91 & 0.47 & \textbf{0.85} & 0.71 & 0.83 & 0.78 \\
InternVL-3 & 0.98 & 0.39 & 0.80 & 0.41 & 0.71 & 0.72 \\
Qwen2.5-VL-32B &\textbf{ 0.99} & \textbf{0.51} & 0.83 & 0.69 & 0.83 & 0.77 \\
Qwen2.5-VL-7B-LoRA & 0.85 & 0.46 & 0.82 & \textbf{0.75} & \textbf{0.87} & \textbf{0.80} \\
\hline
\multicolumn{7}{|c|}{\textbf{Videos}} \\
\hline
Qwen2.5-VL-7B & 0.91 & 0.27 & 0.76 & 0.41 & 0.71 & 0.65 \\
MIMO-SFT-7B & 0.95 & 0.33 & 0.79 & 0.39 & 0.67 & 0.68 \\
InternVL-3 & 0.94 & 0.34 & 0.68 & 0.33 & 0.60 & 0.55 \\
Qwen2.5-VL-32B & 0.97 & 0.34 & 0.70 & 0.49 & 0.57 & 0.62 \\
\hline
\end{tabular}
\label{tab:overall_results}
\end{center}
\end{table*}
\subsubsection{Completeness and Faithfulness Metrics}

Through KIU matching, we compute two fundamental metrics:

\textbf{Completeness (Recall)} measures the proportion of ground truth information units successfully matched:
\begin{equation}
\text{Completeness} = \frac{\text{GT KIUs with matches}}{\text{Total GT KIUs}}
\end{equation}

\textbf{Faithfulness (Precision)} measures the proportion of predicted information units successfully matched:
\begin{equation}
\text{Faithfulness} = \frac{\text{Predicted KIUs with matches}}{\text{Total predicted KIUs}}
\end{equation}

This combination provides comprehensive assessment distinguishing between models offering thorough but potentially inaccurate descriptions versus conservative but highly reliable outputs---a crucial distinction for enterprise deployment decisions where applications may prioritize completeness versus accuracy depending on downstream risk tolerance.

\medskip

Through this comprehensive evaluation framework, \textsc{VLM-in-the-Wild} provides the first systematic, enterprise-grounded assessment methodology for generative Vision-Language Models. By addressing the fundamental challenges of unordered, variable-granularity outputs while maintaining the precision required for business decision-making, our framework enables confident evaluation and deployment of VLMs in real-world enterprise applications.

\section{Experiments and Results}

We evaluate ViLD on three leading open-source 7B-scale Vision-Language Models—Qwen2.5-VL-7B\cite{qwen}, MIMO-SFT-7B\cite{mimo}, and InternVL-3-7B\cite{internvl}—using identical enterprise tasks and unified metrics. To isolate the effects of model scaling and domain adaptation, we additionally assess the larger Qwen2.5-VL-32B and a domain-adapted Qwen2.5-VL-7B (via LoRA fine-tuning on 2,000 curated enterprise samples; hyperparameters and training protocol detailed in Appendix A). This ablation study allows us to disentangle the impact of scale and targeted fine-tuning on enterprise scenario performance. Full results are reported in Table~\ref{tab:overall_results}, which presents the primary metrics for cross-model comparison; comprehensive task-specific metrics for all enterprise tasks are provided in the Appendix B.
\subsection{Main Findings}

\begin{itemize}
\item \textbf{7B Model Landscape:} Among 7B-scale models, MIMO-SFT-7B delivers the best overall balance of accuracy and reliability, matching or surpassing both Qwen2.5-VL-7B and InternVL-3-7B in key tasks. Notably, InternVL-3-7B achieves a high reliability score, though its output require extra parsing due to schema inconsistencies.
\item \textbf{Few-Shot Domain Adaptation:} LoRA fine-tuning on a modest dataset enables Qwen2.5-VL-7B to rival and, in logo, OCR, and media description tasks, even slightly outperform the much larger 32B baseline. This indicates strong specialization potential for targeted business domains.
\item \textbf{Reliability–Accuracy Trade-off:} Fine-tuning boosts specialized task scores but sharply lowers overall reliability, most acutely on video inputs. This exposes a critical risk for enterprise deployment if robustness safeguards are not in place.
\item \textbf{Benefits of Model Scaling:} Qwen2.5-VL-32B consistently achieves the highest task accuracy and reliability, underscoring the advantages of scale for robust enterprise performance.
\item \textbf{Persistent Video Gap:} All models, regardless of scale or adaptation, exhibit pronounced performance degradation on video-based tasks versus images, highlighting ongoing challenges in open multimodal modeling.
\end{itemize}
Taken together, these findings illustrate the essential trade-offs between scale, task specialization, and reliability in practical enterprise scenarios—gaps that ViLD's comprehensive metrics are designed to expose.
\section{Conclusion}
In this work, we present \textbf{VLM-in-the-Wild (ViLD)}, a holistic framework for evaluating Vision-Language Models in authentic enterprise contexts. Our core contributions are:
\begin{itemize}
\item A business-grounded task suite targeting real enterprise needs—from OCR and logo detection to content moderation and scene analysis.
\item A generalized spatial-temporal grid methodology, enabling consistent localization evaluation for generative and structured models alike.
\item The BlockWeaver algorithm for robust, format-agnostic OCR evaluation across diverse VLM outputs.
\item Comprehensive, empirical benchmarking of leading open-source VLMs and domain-adapted models, revealing actionable deployment insights.
\end{itemize}
Our results illuminate both the impressive progress and persistent reliability challenges facing current open VLMs—especially as they move from image-only academic benchmarks to demanding, high-stakes enterprise applications. We envisage that the ViLD framework and dataset foster further community progress toward production-ready, robust, and business-aligned vision-language intelligence.

\appendix
\subsection{Fine-tuning Details}
We fine-tuned the Qwen2.5-VL-7B-Instruct model using Low-Rank Adaptation (LoRA) on our curated dataset of 2,000 video samples. The fine-tuning process was designed to adapt the pre-trained vision-language model to our specific domain while maintaining computational efficiency and preserving the model's general capabilities.

\textbf{Model Architecture and Adaptation Strategy.} We employed LoRA for parameter-efficient fine-tuning while keeping the core vision tower, language model, and multimodal fusion layers frozen. Vision LoRA was specifically enabled to adapt visual encoder components, allowing the model to learn domain-specific visual representations. LoRA modules were strategically excluded from the language model head and embedding tokens to preserve pre-trained linguistic representations and prevent catastrophic forgetting. The complete fine-tuning configuration is detailed in Table~\ref{tab:finetuning_config}, where we set the LoRA rank and alpha both to 64 to balance adaptation capacity with parameter efficiency. A conservative dropout rate of 0.05 was applied to prevent overfitting given our relatively small dataset size.

\textbf{Optimization and Training Regime.} Our optimization strategy utilized a learning rate of \(2 \times 10^{-4}\) with cosine annealing scheduler and a warmup ratio of 0.03 to ensure stable training dynamics. Weight decay was set to 0.1 to provide regularization. We employed a global batch size of 64 implemented through gradient accumulation over 16 steps, enabling effective training despite memory constraints. The model was trained for a single epoch to prevent overfitting while ensuring sufficient exposure to the training data, as shown in the configuration parameters in Table~\ref{tab:finetuning_config}.

\textbf{Vision Processing Configuration.} The vision processing pipeline was configured to handle both image and video inputs with adaptive resolution settings, as detailed in Table~\ref{tab:vision_config}. For images, the resolution ranged from 224×224 pixels minimum to 1280×28×28 pixels maximum, accommodating various input sizes while maintaining computational efficiency. Video processing utilized a frame sampling rate of 1.0 fps with resolutions spanning from 16×28×28 pixels to 602,112 pixels maximum. All computations were performed using bf16 precision to optimize memory usage while maintaining numerical stability.

\textbf{Training Infrastructure and Optimization.} The training was conducted on a single NVIDIA A100 GPU utilizing DeepSpeed ZeRO Stage 2 optimization for enhanced memory efficiency and gradient synchronization. We implemented gradient checkpointing to trade computation for memory, enabling training of larger effective batch sizes. Mixed precision training with bf16 was employed to optimize computational resources while maintaining numerical stability throughout the training process. The data loading pipeline utilized lazy preprocessing with 8 dataloader workers to ensure efficient data handling and minimize I/O bottlenecks.

\begin{figure}[h]
    \centering
    \includegraphics[width=0.5\textwidth]{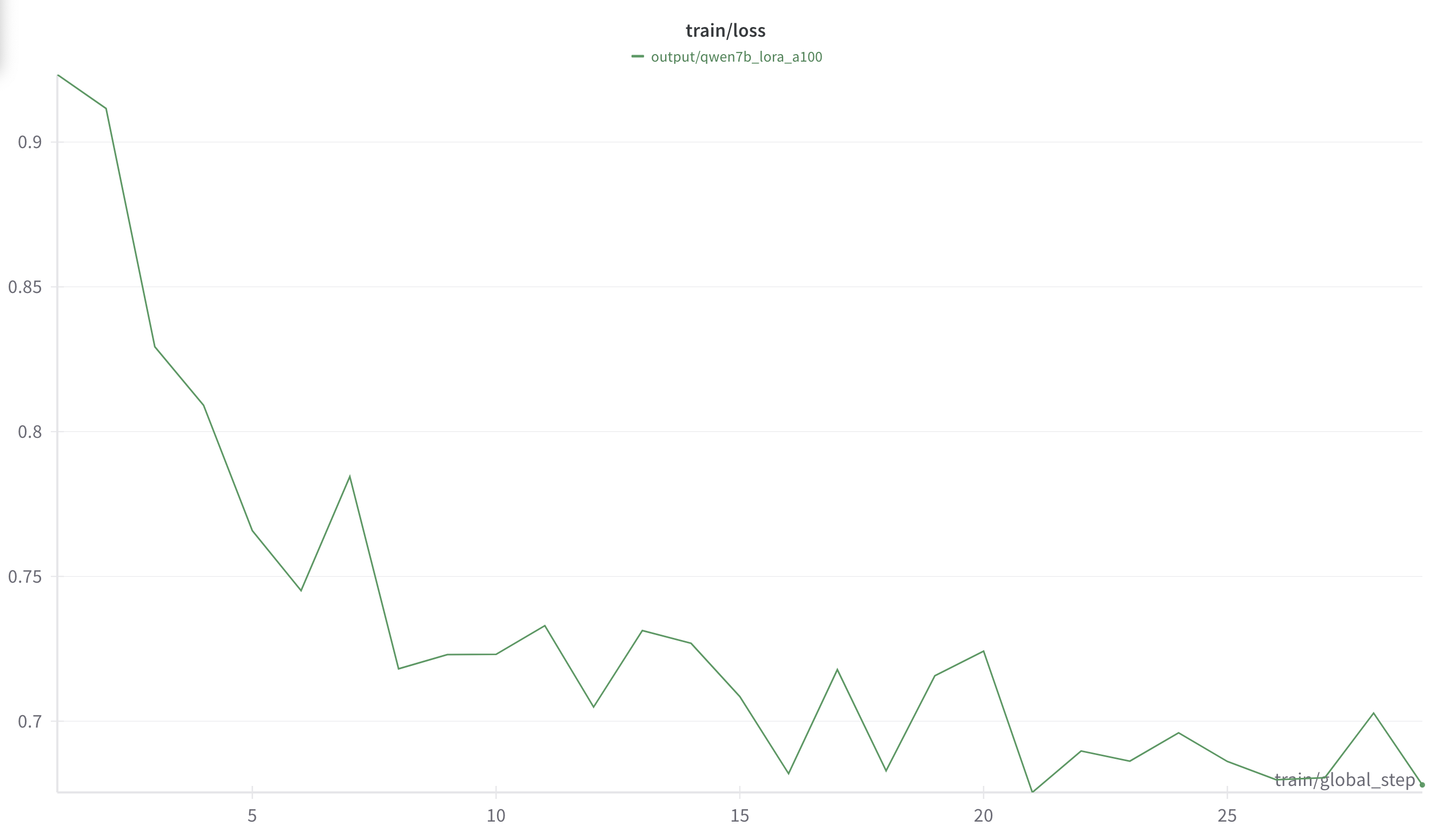}
    \caption{Training loss progression during LoRA fine-tuning of Qwen2.5-VL-7B on our dataset. The model shows stable convergence with consistent loss reduction over training steps, indicating effective domain adaptation without overfitting.}
    \label{fig:training_loss}
\end{figure}

\textbf{Training Dynamics and Convergence.} Figure~\ref{fig:training_loss} illustrates the training loss progression throughout the fine-tuning process. The plot shows a clear downward trend, indicating that the model is effectively learning from the training data.

\begin{table}[htbp]
\caption{Fine-tuning Configuration Parameters}
\begin{center}
\begin{tabular}{|l|l|c|}
\hline
\textbf{Parameter} & \textbf{Category} & \textbf{Value} \\
\hline
Base Model & Architecture & Qwen2.5-VL-7B-Instruct \\
\hline
LoRA Rank & Adaptation & 64 \\
\hline
LoRA Alpha & Adaptation & 64 \\
\hline
LoRA Dropout & Adaptation & 0.05 \\
\hline
Learning Rate & Optimization & \(2 \times 10^{-4}\) \\
\hline
Weight Decay & Optimization & 0.1 \\
\hline
Warmup Ratio & Optimization & 0.03 \\
\hline
LR Scheduler & Optimization & Cosine \\
\hline
Global Batch Size & Training & 64 \\
\hline
Gradient Accumulation & Training & 16 steps \\
\hline
Training Epochs & Training & 1 \\
\hline
\end{tabular}
\label{tab:finetuning_config}
\end{center}
\end{table}

\begin{table}[htbp]
\caption{Vision Processing Configuration}
\begin{center}
\begin{tabular}{|l|l|c|}
\hline
\textbf{Parameter} & \textbf{Modality} & \textbf{Value} \\
\hline
Min Resolution & Image & 224×224 pixels \\
\hline
Max Resolution & Image & 1280×28×28 pixels \\
\hline
Min Resolution & Video & 16×28×28 pixels \\
\hline
Max Resolution & Video & 602,112 pixels \\
\hline
Frame Sampling Rate & Video & 1.0 fps \\
\hline
Precision & Both & bf16 \\
\hline
\end{tabular}
\label{tab:vision_config}
\end{center}
\end{table}


\subsection{Detailed Experimental Results}

This appendix presents comprehensive evaluation results across all tasks and models. For each table, we report both global aggregate metrics (computed over the entire test set) and per-image average metrics (averaged across individual samples). All metrics are reported on a scale of 0-1 unless otherwise specified.

\begin{table*}[htbp]
\caption{Object Detection Performance Metrics. We evaluate detection accuracy using F1 scores computed via two complementary approaches: LLM-based semantic matching and embedding similarity matching.}
\begin{center}
\begin{tabular}{|l|c|c|c|c|}
\hline
\textbf{Model Configuration} & \multicolumn{2}{c|}{\textbf{Global Aggregate}} & \multicolumn{2}{c|}{\textbf{Per-Sample Average}} \\
\cline{2-5} 
 & \textbf{Semantic F1} & \textbf{Embedding F1} & \textbf{Semantic F1} & \textbf{Embedding F1} \\
\hline
Qwen-7B LoRA (Image) & 0.4695 & 0.4831 & 0.3952 & 0.4347 \\
Qwen-7B LoRA (Video) & 0.3574 & 0.3560 & 0.3006 & 0.3196 \\
Qwen-32B (Image) & 0.5150 & 0.4460 & 0.4384 & 0.3973 \\
Qwen-32B (Video) & 0.3453 & 0.2785 & 0.2884 & 0.2500 \\
Qwen-7B Base (Image) & 0.3980 & 0.2209 & 0.3266 & 0.1687 \\
Qwen-7B Base (Video) & 0.2780 & 0.1142 & 0.2431 & 0.0925 \\
InternVL3-8B (Image) & 0.3948 & 0.1722 & 0.3159 & 0.1285 \\
InternVL3-8B (Video) & 0.3487 & 0.1689 & 0.2806 & 0.1316 \\
MIMO-SFT (Image) & 0.4746 & 0.4261 & 0.3825 & 0.3704 \\
MIMO-SFT (Video) & 0.3310 & 0.2523 & 0.2630 & 0.2119 \\
\hline
\end{tabular}
\label{tab:object_detection}
\end{center}
\end{table*}

\begin{table*}[htbp]
\caption{Human Analysis Performance Metrics. Metrics include LLM Scores (0-1 Scale) for activity recognition and visual description quality, demographic attribute accuracy (age, expression, face detection), spatial localization (IoU), and temporal consistency (for video inputs).}
\begin{center}
\resizebox{\textwidth}{!}{%
\begin{tabular}{|l|c|c|c|c|c|c|c|c|c|}
\hline
\textbf{Model Configuration} & \textbf{Activity} & \textbf{Description} & \textbf{Age} & \textbf{Expression} & \textbf{Face Det.} & \textbf{Spatial IoU} & \textbf{Temporal IoU} & \textbf{Global F1} & \textbf{Avg. F1} \\
 & \textbf{Accuracy} & \textbf{Quality} & \textbf{Accuracy} & \textbf{Accuracy} & \textbf{Accuracy} & & & & \\
\hline
Qwen-7B LoRA (Image)    & 0.6915 & 0.7088 & 0.9027 & 0.7687 & 0.7882 & 0.5875 & 0.9968 & 0.7509 & 0.8258 \\
Qwen-7B LoRA (Video)    & 0.6170 & 0.6483 & 0.8944 & 0.5628 & 0.7974 & 0.4594 & 0.7493 & 0.6421 & 0.7504 \\
Qwen-32B (Image)        & 0.6984 & 0.6679 & 0.8838 & 0.6991 & 0.7692 & 0.3845 & 1.0000 & 0.7640 & 0.8394 \\
Qwen-32B (Video)        & 0.6350 & 0.5741 & 0.8957 & 0.6325 & 0.8008 & 0.3814 & 0.6444 & 0.5726 & 0.7013 \\
Qwen-7B Base (Image)    & 0.6539 & 0.5725 & 0.7829 & 0.6377 & 0.7175 & 0.3014 & 0.9990 & 0.6840 & 0.7832 \\
Qwen-7B Base (Video)    & 0.6222 & 0.5425 & 0.8296 & 0.5058 & 0.7788 & 0.3581 & 0.4242 & 0.6118 & 0.7635 \\
InternVL3-8B (Image)    & 0.6639 & 0.5969 & 0.7974 & 0.7297 & 0.6856 & 0.3032 & 0.9791 & 0.7255 & 0.8024 \\
InternVL3-8B (Video)    & 0.6535 & 0.5707 & 0.1264 & 0.3342 & 0.3359 & 0.3524 & 0.3861 & 0.5187 & 0.6812 \\
MIMO-SFT (Image)        & 0.6691 & 0.6991 & 0.8596 & 0.7733 & 0.7993 & 0.3773 & 0.9933 & 0.7867 & 0.8538 \\
MIMO-SFT (Video)        & 0.6337 & 0.6651 & 0.8199 & 0.5395 & 0.8045 & 0.3929 & 0.7020 & 0.6429 & 0.7899 \\
\hline
\end{tabular}
}
\label{tab:human_analysis}
\end{center}
\end{table*}

\begin{table*}[htbp]
\caption{Logo Detection Performance Metrics. We report F1 scores for logo identification and Jaccard similarity indices for spatial (position) localization accuracy.}
\begin{center}
\resizebox{0.6\textwidth}{!}{%
\begin{tabular}{|l|c|c|c|c|}
\hline
\textbf{Model Configuration} & \multicolumn{2}{c|}{\textbf{Global Metrics}} & \multicolumn{2}{c|}{\textbf{Per-Sample Average}} \\
\cline{2-5}
 & \textbf{F1 Score} & \textbf{Spatial IoU} & \textbf{F1 Score} & \textbf{Spatial IoU} \\
\hline
Qwen-7B LoRA (Image)    & 0.7552 & 0.5891 & 0.6770 & 0.6377 \\
Qwen-7B LoRA (Video)    & 0.5804 & 0.3919 & 0.5669 & 0.4033 \\
Qwen-32B (Image)        & 0.6920 & 0.4835 & 0.5958 & 0.5397 \\
Qwen-7B Base (Image)    & 0.5493 & 0.4168 & 0.4925 & 0.4799 \\
Qwen-7B Base (Video)    & 0.4158 & 0.3836 & 0.3440 & 0.4188 \\
InternVL3-8B (Image)    & 0.5420 & 0.4140 & 0.4350 & 0.4503 \\
InternVL3-8B (Video)    & 0.4978 & 0.3334 & 0.5330 & 0.3473 \\
MIMO-SFT (Image)        & 0.7139 & 0.5223 & 0.6308 & 0.5637 \\
MIMO-SFT (Video)        & 0.3901 & 0.3803 & 0.2780 & 0.4089 \\
\hline
\end{tabular}
}
\label{tab:logo_detection}
\end{center}
\end{table*}

\begin{table*}[htbp]
\caption{Media Analysis Performance Metrics. We report exact match rate for media quality, Jaccard similarity for dominant color detection, LLM scores (1–5 scale) for camera perspective and scene detection, and F1 scores for detailed description quality.}
\begin{center}
\begin{tabular}{|l|c|c|c|c|c|}
\hline
\textbf{Model Configuration} & \textbf{Media Quality} & \textbf{Colors (Jaccard)} & \textbf{Camera Perspective} & \textbf{Scene Detection} & \textbf{Description (F1)} \\
\hline
Qwen-7B LoRA (Image)      & 0.9521 & 0.7168 & 3.4507 & 3.5056 & 0.8003 \\
Qwen-7B LoRA (Video)      & 0.7319 & 0.6511 & 3.2250 & 3.2797 & 0.6483 \\
Qwen-32B (Image)          & 0.9453 & 0.6827 & 3.4170 & 3.2518 & 0.7783 \\
Qwen-32B (Video)          & 0.6648 & 0.6318 & 3.0700 & 3.1338 & 0.6236 \\
Qwen-7B Base (Image)      & 0.9480 & 0.6240 & 3.1526 & 3.0585 & 0.7071 \\
Qwen-7B Base (Video)      & 0.7273 & 0.5741 & 3.1181 & 3.0836 & 0.6540 \\
InternVL3-8B (Image)      & 0.9497 & 0.6079 & 3.0129 & 2.9124 & 0.7253 \\
InternVL3-8B (Video)      & 0.7038 & 0.5878 & 2.9203 & 3.1008 & 0.6607 \\
MIMO-SFT (Image)          & 0.9313 & 0.6522 & 3.3179 & 3.5037 & 0.7876 \\
MIMO-SFT (Video)          & 0.7654 & 0.5950 & 3.0514 & 3.2334 & 0.6853 \\
\hline
\end{tabular}
\label{tab:media_analysis}
\end{center}
\end{table*}

\begin{table*}[htbp]
\caption{Optical Character Recognition (OCR) Performance Metrics. We report both character-level and word-level metrics. CER: Character Error Rate, WER: Word Error Rate (lower is better). F1 scores measure recognition accuracy at character and word granularities.}
\begin{center}
\resizebox{\textwidth}{!}{%
\begin{tabular}{|l|c|c|c|c|c|c|c|c|}
\hline
\textbf{Model Configuration} & \multicolumn{4}{c|}{\textbf{Per-Sample Average}} & \multicolumn{4}{c|}{\textbf{Corpus-Level}} \\
\cline{2-9}
 & \textbf{CER↓} & \textbf{WER↓} & \textbf{Char F1} & \textbf{Word F1} & \textbf{CER↓} & \textbf{WER↓} & \textbf{Char F1} & \textbf{Word F1} \\
\hline
Qwen-7B LoRA (Image)    & 0.1982 & 0.3244 & 0.8768 & 0.8244 & 0.2103 & 0.3115 & 0.9082 & 0.8853 \\
Qwen-7B LoRA (Video)    & 0.5835 & 0.7019 & 0.7209 & 0.6435 & 0.5728 & 0.6778 & 0.6030 & 0.5086 \\
Qwen-32B (Image)        & 0.2267 & 0.3371 & 0.8325 & 0.7733 & 0.2099 & 0.3138 & 0.8425 & 0.8066 \\
Qwen-32B (Video)        & 0.6409 & 0.7983 & 0.5743 & 0.4813 & 0.6319 & 0.7631 & 0.4003 & 0.2651 \\
Qwen-7B Base (Image)    & 0.2691 & 0.3768 & 0.7546 & 0.6935 & 0.2332 & 0.3323 & 0.7550 & 0.7196 \\
Qwen-7B Base (Video)    & 0.2316 & 0.3428 & 0.7194 & 0.6881 & 0.2461 & 0.3333 & 0.6740 & 0.6481 \\
InternVL3-8B (Image)    & 0.3718 & 0.4594 & 0.7162 & 0.6436 & 0.3083 & 0.4149 & 0.6471 & 0.6009 \\
InternVL3-8B (Video)    & 0.4332 & 0.6068 & 0.6034 & 0.5362 & 0.3879 & 0.5163 & 0.5008 & 0.4385 \\
MIMO-SFT (Image)        & 0.2169 & 0.3658 & 0.8318 & 0.7625 & 0.2020 & 0.3341 & 0.8502 & 0.8145 \\
MIMO-SFT (Video)        & 0.2506 & 0.3979 & 0.6792 & 0.6225 & 0.2700 & 0.3785 & 0.6265 & 0.5846 \\
\hline
\end{tabular}
}
\label{tab:ocr_results}
\end{center}
\end{table*}

\end{document}